\documentclass[review]{elsarticle}

\usepackage{lineno,hyperref}
\usepackage[tight,footnotesize]{subfigure}

\usepackage{longtable}
\usepackage{booktabs}
\usepackage{url}

\usepackage{multirow}
\usepackage{amsmath}

\let\oldequation\equation
\let\oldendequation\endequation

\renewenvironment{equation}
  {\linenomathNonumbers\oldequation}
  {\oldendequation\endlinenomath}

\journal{Journal of \LaTeX\ Templates}

\begin{document}

\begin{frontmatter}

\title{An Evaluation of Machine Learning and Deep Learning Models for Drought Prediction using Weather Data}

\author{Weiwei Jiang\corref{mycorrespondingauthor}}
\address{Department of Electronic Engineering, Tsinghua University, Beijing, 100084, China}
\cortext[mycorrespondingauthor]{Corresponding author. E-mail address: jwwthu@gmail.com}

\author{Jiayun Luo}
\address{Department of Statistics, University of California-Los Angeles, Los Angeles, 90024, USA}

\begin{abstract}
Drought is a serious natural disaster that has a long duration and a wide range of influence. To decrease the drought-caused losses, drought prediction is the basis of making the corresponding drought prevention and disaster reduction measures. While this problem has been studied in the literature, it remains unknown whether drought can be precisely predicted or not with machine learning models using weather data. To answer this question, a real-world public dataset is leveraged in this study and different drought levels are predicted using the last 90 days of 18 meteorological indicators as the predictors. In a comprehensive approach, 16 machine learning models and 16 deep learning models are evaluated and compared. The results show no single model can achieve the best performance for all evaluation metrics simultaneously, which indicates the drought prediction problem is still challenging. As benchmarks for further studies, the code and results are publicly available in a Github repository.
\end{abstract}


\end{frontmatter}


\section{Introduction}
\label{sec:intro}
Drought is a natural phenomenon that occurs when the precipitation is significantly lower than the normal level for a period of time. It develops slowly over time, has complex causes, and has a long duration and a wide range of influence. Drought has caused major damage to the natural environment, human life and social economy. It is one of the most widespread, common and catastrophic natural disasters in the world today, and the losses caused by drought are far greater than other meteorological disasters. Take China as an example. On average, at least one severe drought occurs in China every year, and drought-related losses account for 30\% of the country's total disaster losses. From a global perspective, in the occurrence of natural disasters, 22\% of the economic losses and 33\% of the affected population can be attributed to drought.

Drought is not only affected by natural factors such as precipitation, temperature and evapotranspiration, but also closely related to human activities such as over-farming, over-irrigation, deforestation, and over-exploitation of available water resources. In recent years, with the impact of global warming and rapid social and economic development, the frequency, intensity, and scope of droughts have increased, and the damage caused by them has become more serious. Drought has attracted widespread attention from the academia, the government and the public globally. Accurate drought prediction plays an important role in planning and managing regional water resources and mitigating the harmful effects of drought. It also provides scientific basis for drought monitoring, early warning and risk assessment, and helps relevant departments to make decisions in the corresponding drought prevention and disaster reduction measures.

To quantify the drought levels, different indices are used in previous studies, including Standard Precipitation Evaporation Index (SPEI)~\cite{tian2018agricultural, dikshit2021improved, mehr2021drought, mokhtar2021estimation, dikshit2021long, khan2020prediction, zhang2019meteorological}, Standardized Precipitation Index (SPI)~\cite{mehr2021drought, xu2018evaluation, zhu2021internal}, Groundwater Resource Index (GRI)~\cite{banadkooki2021multi} and Effective Drought Index (EDI)~\cite{malik2021prediction}. While these indices are continuous values, a discrete set of drought levels can be set by dividing different value ranges. Both formulations are widely used in the literature, i.e., the regression formulation when an index is predicted and the classification formulation when the drought level is predicted. The latter formulation is adopted in this study.

As effective tools, various machine learning models have already been applied in the environmental research. Machine learning aims to learn a mapping relationship between the input features and targets. Unlike weather models, machine learning does not depend on the domain knowledge or physical intrinsic mechanisms. However, machine learning models usually require a lot of data to work, which is not a serious problem in the environmental area with more and more observations accumulated in past decades. As a specific type of machine learning, deep learning is represented by various deep neural networks, which have been extremely successfully in the past decade for a series of problems, e.g., image recognition, time series prediction, etc~\cite{jiang2018geospatial, jiang2020applications, jiang2020edge}. Their effectiveness for drought prediction would also be evaluated in this study, along with the traditional machine learning models.

The relevant studies of applying machine learning techniques for drought prediction would be discussed in Section~\ref{sec:related}. However, there are still some shortcomings in these studies. In most studies, only two or three models are used, without giving a comprehensive evaluation or consideration for other possible alternatives. Since many deep learning models are proposed in recent years, they have not been applied in hydrological studies yet.

In this study, we aim to give a comprehensive evaluation that covers most of the popular machine learning and deep learning models for drought prediction as a case study. We also aim to incorporate some of the latest progress from the deep learning field, e.g., Time Series Transformer, which has not been considered for this specific task in the current literature as far as the authors know. To summarize, a total of 16 machine learning models and 16 deep learning models are used in this study. The machine learning models include Logistic Regression (LR)~\cite{bishop2006pattern}, k Nearest Neighbor (KNN)~\cite{altman1992introduction}, Naivers Bayes (NB)~\cite{friedman1997bayesian}, Support Vector Machine (SVM)~\cite{cortes1995support} with the linear and Radial Basis Function (RBF) kernels, Linear Discriminant Analysis (LDA)~\cite{tharwat2016linear}, Quadratic Discriminant Analysis (QDA)~\cite{tharwat2016linear}, Ridge Classifier (Ridge)~\cite{saunders1998ridge}, Decision Tree (DT)~\cite{breiman1984classification}, Random Forest (RF)~\cite{breiman2001random}, Extra Trees (ET)~\cite{geurts2006extremely}, AdaBoost Classifier (AdaBoost)~\cite{freund1997decision}, Gradient Boosting Classifier (GB)~\cite{friedman2001greedy}, Extreme Gradient Boosting (XGBoost)~\cite{chen2016xgboost}, Light Gradient Boosting (LightGBM)~\cite{ke2017lightgbm}, and CatBoost Classifier~\cite{prokhorenkova2018catboost}. The deep learning models include Multilayer Perceptron (MLP)~\cite{fawaz2019deep}, Recurrent Neural Network (RNN)~\cite{rumelhart1986learning}, Gated Recurrent Unit (GRU)~\cite{cho2014learning}, Long Short-Term Memory (LSTM)~\cite{hochreiter1997long}, Fully Convolutional Network (FCN)~\cite{wang2017time}, Residual Neural Network (ResNet)~\cite{wang2017time}, Residual Convolutional Neural Network (ResCNN)~\cite{wang2017time}, Temporal Convolutional Networks (TCN)~\cite{bai2018empirical}, InceptionTime~\cite{fawaz2020inceptiontime}, XceptionTime~\cite{rahimian2020xceptiontime}, Omni-Scale 1D-CNN (OS-CNN)~\cite{tang2020rethinking}, XCM~\cite{fauvel2020xcm}, RNN\_FCN~\cite{karim2017lstm}, GRU\_FCN~\cite{karim2017lstm}, LSTM\_FCN~\cite{karim2017lstm}, and Time Series Transformer (TST)~\cite{zerveas2020transformer}.

In this study, a specific dataset is leveraged to evaluate the performance of different machine learning and deep learning models. This public dataset collects 18 meteorological indicators across the United States from 2000 to 2020, which contain wind speed, temperature, surface pressure, humidity, precipitation, etc. They are used as the predictors in the machine learning and deep learning models, to predict five different drought levels, which are defined on well-known drought indices, e.g., SPI. While this study is based on meteorological data which are also used in many previous studies, other data sources can also be used for drought prediction. For example, without using meteorological data, satellite image and topography data are used to predict the severe drought area with random forest~\cite{park2019prediction}.

Five classification evaluation metrics are used for comparing the performance, namely, accuracy, precision, recall, F1 score and Matthews Correlation Coefficient (MCC). Considering the imbalanced classes, the macro average values are used for these metrics. Our experiments demonstrate that no single model can achieve the best performance for all evaluation metrics simultaneously. Overall, deep learning models perform better than machine learning models. Among machine learning models, XGBoost achieves the highest accuracy and SVM with the RBF kernel achieves the highest F1 score and MCC. Among deep learning models, GRU achieves the highest accuracy, LSTM achieves the highest F1 score, and XceptionTime achieves the highest MCC.

Our contributions in this study are summarized as follows:
\begin{itemize}
    \item \textit{Comprehensive Evaluation}. For drought level prediction using weather data, a comprehensive evaluation of state-of-the-art machine learning and deep learning models are conducted in this study, based on a real-world dataset collected in US spanning from 2000 to 2020.
    \item \textit{Benchmarks}. Besides the data which is publicly available in the Kaggle website~\footnote{Source: \url{https://www.kaggle.com/cdminix/us-drought-meteorological-data}}, the code and results are also made publicly available in a Github repository~\footnote{\url{https://github.com/jwwthu/DL4Climate/tree/main/DroughtPrediction}}, for any further studies as benchmarks.
    \item \textit{Future Directions}. Besides the findings from our experiments, two possible future directions are also pointed out for inspiring future studies.
\end{itemize}

The remainder of this paper is organized as follows. In Section~\ref{sec:related}, relevant research studies are reviewed. In Section~\ref{sec:problem}, the prediction problem and dataset are described. In Section~\ref{sec:methods}, both the machine learning and deep learning models are introduced. In Section~\ref{sec:experiments}, the experiments and results are discussed. In Section~\ref{sec:conclusion}, the conclusion is drawn and the future research directions are pointed out.

\section{Related Work}
\label{sec:related}
In the literature, machine learning models are already used, with different input data and prediction targets. Compared with statistical models, machine learning models show a better performance in a series of studies. In this section, only those using machine learning methods are discussed. More related work can be found in a recent survey~\cite{hao2018seasonal}.

While machine learning and deep learning models are frequently evaluated and compared in various areas, there is no certain conclusion that which model always wins. Without comparing deep learning models, machine learning models are compared with statistical models or compared with each other. For Xiangjiang River basin, the relation between soil moisture and drought is quantified in~\cite{tian2018agricultural}. It is found that among the climate factors that influence the agriculture drought, the Ridge Point of western Pacific subtropical high (WPSH) is the key factor. Then the Support Vector Regression (SVR) model is proposed for the agricultural drought prediction. Compared with solely using drought index as input, the SVR model incorporating climate indices improves the prediction accuracy by 5.1\% in testing. For drought prediction in China, a series of statistical, dynamic and hybrid models are evaluated in~\cite{xu2018evaluation}, and the results show that the ensemble streamflow prediction (ESP) method and wavelet machine learning models outperform other statistical models in forecasting SPI in six months (SPI6). For one-month ahead prediction of SPI in Ankara province and SPEI in central Antalya region, three types of machine learning models, namely, decision tree, genetic programming, and gradient boosting decision tree, are evaluated and compared in~\cite{mehr2021drought}. Three classes of wet, normal, and dry events are used to model the drought situation. And the results demonstrate the superiority of the proposed gradient boosting decision tree.

When traditional machine learning models are compared with deep learning models, the winner is hard to tell. For a case study in the Tibetan Plateau, China, for the period of 1980-2019, four machine learning models, namely Random Forest (RF), the Extreme Gradient Boost (XGB), the Convolutional neural network (CNN) and the Long-term short memory (LSTM), are developed and evaluated in~\cite{mokhtar2021estimation}, for the prediction of SPEI3 and SPEI6. Various combinations of climate variables are used as model inputs under different scenarios. Their results also demonstrate that there is no single winner model for all the scenarios, in which XGB and RF models are relatively better than the other two models. Three machine learning methods, namely, SVM, ANN and KNN, are developed in~\cite{khan2020prediction} to predict the drought level over Pakistan, which is classified into three types as moderate, severe and extreme. To choose suitable input features, a novel feature selection named Recursive Feature Elimination (RFE) is leveraged and the results show ANN and KNN-based models perform better than SVM. Instead of using a single weather station, data from 32 stations during 1961 to 2016 in Shaanxi province, China are used in~\cite{zhang2019meteorological}. To select input features, a cross-correlation function and a distributed lag nonlinear model (DLNM) is applied. Then two machine learning models are used for predicting SPEI 1-6 months in advance, including XGBoost and ANN. The results show that XGBoost has a higher prediction accuracy.

Over the New South Wales (NSW) region of Australia, the Long Short-Term Memory (LSTM) is used to predict the SPEI at two different time scales (SPEI1, SPEI3) in~\cite{dikshit2021improved}, using hydro-meteorological variables as model inputs. The Climatic Research Unit (CRU) dataset from 1901 to 2018 is used to validate the performance of LSTM, compared with other machine learning models including random forests and artificial neural networks. The results show that LSTM achieves an AUC value of 0.83 and 0.82 for SPEI1 and SPEI3 respectively. Using data spanning from 1901 to 2018, a stacked LSTM model is proposed in~\cite{dikshit2021long} and used to predict SPEI from 1 month to 12 months in the New South Wales region of Australia. The hydro-meteorological and climatic variables are used as model inputs. The experiments demonstrate that the stacked LSTM model can outperform traditional ML models.

Instead of comparing different models, different variants of ANN are compared and used for predicting the Groundwater Resource Index (GRI)-based drought at different timescales (6, 12, and 24 months) in Yazd plain, Iran in~\cite{banadkooki2021multi}. These variants are based on the Salp Swarm Algorithm (SSA), Particle Swarm Optimization (PSO), and Genetic Algorithm (GA). The results show that under certain input scenario, these models would perform better. In a similar approach, different variants of SVR are evaluated in~\cite{malik2021prediction} for Effective Drought Index (EDI) 1 month ahead, at different locations of Uttarakhand State of India. The variants are based on two different optimization algorithms, i.e., Particle Swarm Optimization (PSO) and Harris Hawks Optimization (HHO). The results show that SVR-HHO outperforms SVR-PSO.

Probabilistic drought forecasting is also considered in previous studies. Unlike deterministic forecasting which generates a single value, probabilistic forecasting gives the predictive mean and confidence interval, which could be more helpful if the extreme cases are to be considered in practice. Taking the drought prediction in the three-river headwater region of China as the demonstration, an improved LSTM model is proposed in~\cite{zhu2021internal} for probabilistic drought forecasting. The experiments showed that improved LSTM can forecast drought index and uncertainty information better than other machine learning models, e.g., autoregressive moving average, generalized linear regression, and artificial neural network.

While deep learning models are proven effective in many academic studies, they are less adopted in operational hydrology because of the `black-box' nature of deep neural networks. It would be difficult to explain the decision process of deep learning models for drought classification or prediction. To tackle this problem, many efforts have been made to enhance the interpretability of these models. For example, the decision tree is used in~\cite{vidyarthi2020knowledge} to extract knowledge from the input-output relation of a trained ANN model, to establish simple rules for drought forecasting and monitoring.

\section{Problem and Dataset Description}
\label{sec:problem}
In this study, the drought prediction problem across the US is considered, based on an open Kaggle dataset, which is publicly available~\footnote{Source: \url{https://www.kaggle.com/cdminix/us-drought-meteorological-data}}. The original data comes from NASA POWER Project~\footnote{\url{https://power.larc.nasa.gov/}} and US Drought Monitor~\footnote{\url{https://droughtmonitor.unl.edu/}}, which is a measure of drought across the US manually created by experts using a wide range of data. The problem to investigate in this study is whether droughts could be predicted using only meteorological data, with the help of state-of-the-art machine learning and deep learning methods.

Each data sample in this specific dataset contains the drought level at a specific point in a specific US county, accompanied by the last 90 days of 18 meteorological indicators. These meteorological indicators and their descriptions are shown in Table~\ref{tab:indicators}. They would be used as the input features for various machine learning and deep learning models. These features have already been standardized before being uploaded to the Kaggle website. There are also no missing data in the dataset, thus no further pre-processing techniques are applied in this study.

\begin{table}[!htb]
\centering
\caption{The list of the meteorological indicators used in this study.}
\label{tab:indicators}
\begin{tabular}{ll}
\hline
Indicator & Description \\
\hline
WS10M\_MIN & Minimum Wind Speed at 10 Meters (m/s) \\
QV2M & Specific Humidity at 2 Meters (g/kg) \\
T2M\_RANGE & Temperature Range at 2 Meters (C) \\
WS10M & Wind Speed at 10 Meters (m/s) \\
T2M & Temperature at 2 Meters (C) \\
WS50M\_MIN & Minimum Wind Speed at 50 Meters (m/s) \\
T2M\_MAX & Maximum Temperature at 2 Meters (C) \\
WS50M & Wind Speed at 50 Meters (m/s) \\
TS & Earth Skin Temperature (C) \\
WS50M\_RANGE & Wind Speed Range at 50 Meters (m/s) \\
WS50M\_MAX & Maximum Wind Speed at 50 Meters (m/s) \\
WS10M\_MAX & Maximum Wind Speed at 10 Meters (m/s) \\
WS10M\_RANGE & Wind Speed Range at 10 Meters (m/s) \\
PS & Surface Pressure (kPa) \\
T2MDEW & Dew/Frost Point at 2 Meters (C) \\
T2M\_MIN & Minimum Temperature at 2 Meters (C) \\
T2MWET & Wet Bulb Temperature at 2 Meters (C) \\
PRECTOT & Precipitation (mm day-1) \\
\hline
\end{tabular}
\end{table}

In this study, the drought prediction problem is categorized as a classification problem in the field of machine learning. Different drought classes are shown in Table~\ref{tab:targets}~\footnote{Source: \url{https://droughtmonitor.unl.edu/About/AbouttheData/DroughtClassification.aspx}}, as prediction targets. The value ranges of different drought levels are also shown in Table~\ref{tab:targets}, for the following indices:
\begin{itemize}
    \item Palmer Drought Severity Index (PDSI)~\footnote{\url{http://www.droughtmanagement.info/palmer-drought-severity-index-pdsi/}}, which was developed in the 1960s and is calculated using monthly temperature and precipitation data along with information on the water-holding capacity of soils.
    \item CPC Soil Moisture Model (Percentiles)~\footnote{\url{https://www.cpc.ncep.noaa.gov/products/Soilmst_Monitoring/US/Soilmst/Soilmst.shtml}}, which is estimated by a one-layer hydrological model that takes observed precipitation and temperature and calculates soil moisture, evaporation and runoff.
    \item USGS Weekly Streamflow (Percentiles)~\footnote{\url{https://waterwatch.usgs.gov/}}, which displays maps, graphs, and tables that describe real-time, recent, and past streamflow conditions for the United States, including flood and droughts.
    \item Standardized Precipitation Index (SPI)~\footnote{\url{https://www.droughtmanagement.info/standardized-precipitation-index-spi/}}, which uses historical precipitation records for any location to develop a probability of precipitation that can be computed at any number of timescales. Drought events are indicated when the results of SPI become continuously negative and reach a value of -1. SPI was proposed back to 1992 and was recommended as the main meteorological drought index by World Meteorological Organization (WMO) in 2009.
\end{itemize}

\begin{table}[!htb]
\centering
\caption{The drought classification used in this study.}
\label{tab:targets}
\begin{tabular}{|l|p{2cm}|p{2cm}|p{2cm}|p{2cm}|p{2cm}|}
\hline
\multirow{2}{*}{Class} & \multirow{2}{*}{Description} & \multicolumn{4}{c|}{Range} \\
\cline{3-6}
& & Palmer Drought Severity Index (PDSI) & CPC Soil Moisture Model (Percentiles) & USGS Weekly Streamflow (Percentiles) & Standardized Precipitation Index (SPI) \\
\hline 
No Drought & No Drought & N/A & N/A & N/A & N/A \\
\hline
D0 & Abnormally Dry & -1.0 to -1.9 & 21 to 30 & 21 to 30 & -0.5 to -0.7\\
\hline
D1 & Moderate Drought & -2.0 to -2.9 & 11 to 20 & 11 to 20 & -0.8 to -1.2\\
\hline
D2 & Severe Drought & -3.0 to -3.9 & 6 to 10 & 6 to 10 & -1.3 to -1.5\\
\hline
D3 & Extreme Drought & -4.0 to -4.9 & 3 to 5 & 3 to 5 & -1.6 to -1.9\\
\hline
D4 & Exceptional Drought & -5.0 or less & 0 to 2 & 0 to 2 & -2.0 or less\\
\hline
\end{tabular}
\end{table}

\begin{figure*}[!htb]
\centering
\includegraphics[width=\textwidth]{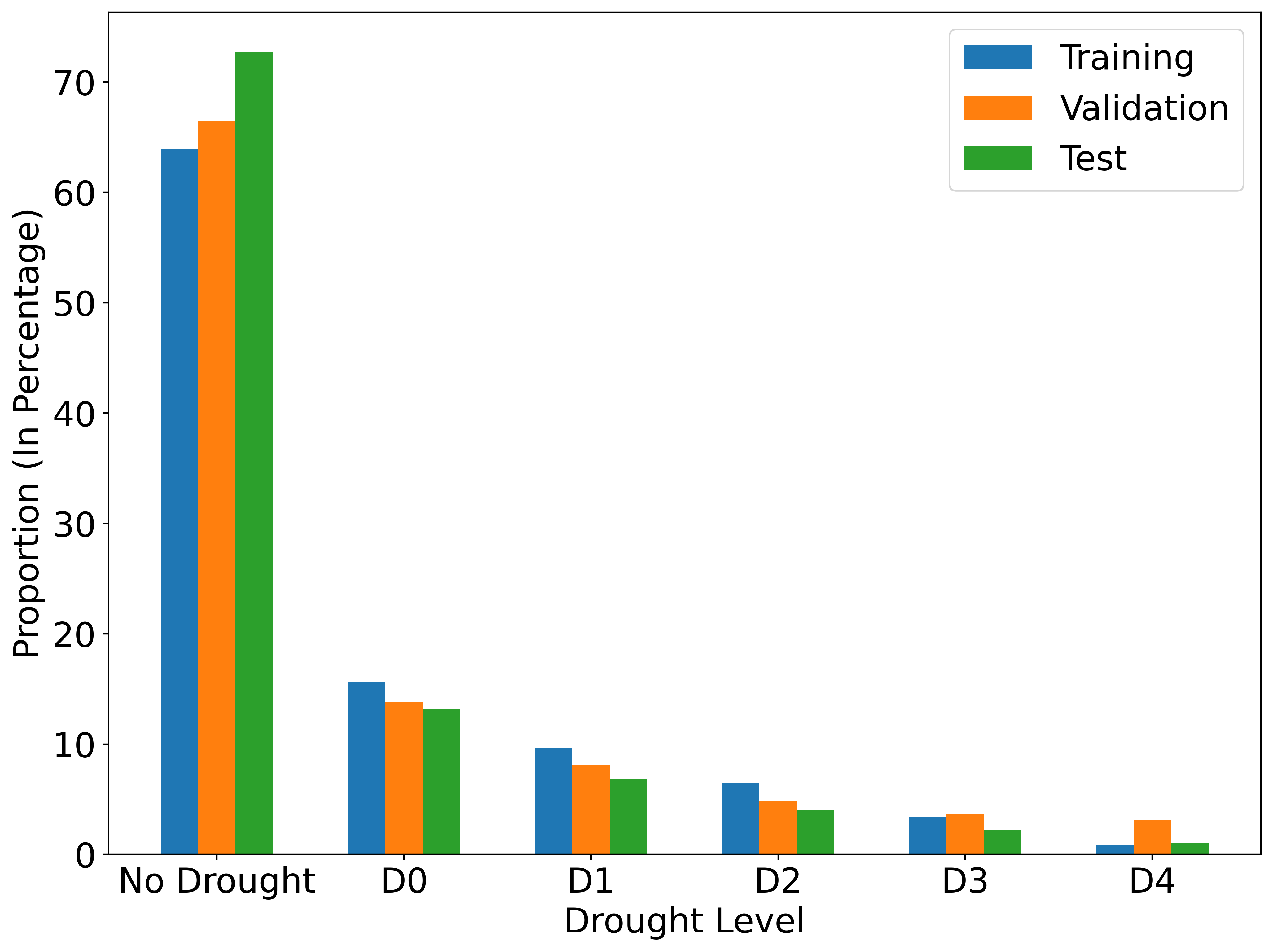}
\caption{The class distribution in three subsets.}
\label{fig:distribution}
\end{figure*}

The whole dataset is split into three subsets, i.e., training set, validation set, and test set, as shown in Table~\ref{tab:split}. The training set is used for updating the model parameters. The validation set is used for choosing the best hyper parameters, which cannot be simply learned from the data. The test set is used to evaluate and compare the performance of different models as the out-of-sample case. The class distributions in three subsets are shown in Figure~\ref{fig:distribution}. As one can tell from Figure~\ref{fig:distribution}, this specific drought prediction problem is challenging because the distribution of classes is highly imbalanced.

\begin{table}[!htb]
\centering
\caption{The data split used in this study.}
\label{tab:split}
\begin{tabular}{|l|l|l|l|}
\hline
Split & Year Range (inclusive) & Data Sample Number & Data Percentage (approximate) \\
\hline
Training & 2000-2009 & 118,024 & 47\% \\
Validation & 2010-2011 & 20,721 & 10\% \\
Test & 2012-2020 & 102,430 & 43\% \\
\hline
\end{tabular}
\end{table}

\section{Methods}
\label{sec:methods}

\subsection{Machine Learning Models}
\subsubsection{Logistic Regression}
Logistic Regression (LR)~\cite{bishop2006pattern} is originally proposed for binary classification tasks, in which the Logistic function is used to map the input $x$ to an output range from 0 to 1, to fit the pattern of binomial data:
\begin{equation}
\sigma(x) = \frac{1} {1 + e^{-x}}
\end{equation}
\noindent As the output of the Logistic function would always be between 0 and 1, it is usually set to be the probability of the output variable belonging to the positive class. Through choosing a threshold, the classification task can be performed by claiming the prediction as one when the output variable is above the threshold, and zero otherwise. For the case of the multi-class classification, the multinomial Logistic regression can be used.

\subsubsection{K Nearest Neighbor}
K Nearest Neighbor (KNN)~\cite{altman1992introduction} is a non-parametric model that can be used for both classification and regression tasks. It is a relatively simple method as it essentially uses K nearest data points in the feature space to determine the predicted value for the data point of interest. Multiple distance measures can be used to determine the proximity between data points, including Minkowski distance, Euclidean distance and Manhattan distance. For the classification task, the majority rule is usually applied and the class to which most of the K nearest data points belong would be the final prediction. For the regression task, the mean value of the K nearest data points would be the final prediction. For determining the optimal K value, the cross validation approach is often used.

\subsubsection{Naivers Bayes}
Naivers Bayes (NB)~\cite{friedman1997bayesian} makes a prediction through estimating posterior probability $p(C|F_1, F_2, ..., F_n)$ from prior probability $p(C)$ and conditional probability $p(F_1, F_2, ..., F_n|C)$. It is called naive because it makes a strong assumption that the features are independent. With this assumption, one can calculate the posterior probability of a specific class $C$ through simply multiplying the prior probability with the probabilities of each feature $F_i$ conditioned on being the class $C$ as shown below:
\begin{equation}
\begin{split}
p(C|F_1, F_2, ..., F_n) & = \frac{p(C) \times p(F_1, F_2, ..., F_n|C)}{p(F_1, F_2, ..., F_n)} \\
& = \frac{1}{Z} p(C) \times p(F_1|C) \times p(F_2|C) \times \dots \times p(F_n|C)
\end{split}
\end{equation}
\noindent where $Z$ is a scaling factor. The class with the highest posterior probability would thus be the final prediction. 

\subsubsection{Support Vector Machine}
Support Vector Machine (SVM)~\cite{cortes1995support} is originally used for binary classification task. Given the data points in the $N$ dimensional space, SVM uses a $N-1$ dimensional threshold or a $N-1$ dimensional hyper-plane to separate two classes, so that the distance between points on the edges for each class to the hyper-plane is maximized. The shortest distance between the observations and the hyper-plane is called the margin $\rho$. Using the formula for calculating the distance between two parallel lines, we get the following formula for the margin:
\begin{equation}
\rho = \frac{2}{\| w \|} 
\end{equation}
\noindent where $w$ is the normal vector to the hyper-plane. Maximizing margin $\rho$ would be equivalent to minimizing $ \frac{1}{2} \| w \|^{2} $, subject to the following constraint:
\begin{equation}
\min_{w, b} \quad \frac{1}{2} \| w \|^{2}
\quad s.t. \quad y_{i} \left( X_i^T \cdot W + b \right) -1 \geq 0, \quad i = 1,2, \cdots, N
\end{equation} 

When allowing misclassifications, the distance between the observations and the hyper-plane is called a soft margin. When using a soft margin to determine the location of a threshold, the hyper-plane is named a soft margin classifier. In the case when two classes overlap or intertwine with each other, classifying data points in their original dimensional space becomes infeasible. In this case, kernel functions, e.g., polynomial kernel or Radial Basis Function (RBF) kernel, can be utilized to map the data points into higher dimensions, so that a classifier can be found after calculating the relationship between each pair of data. Two types of kernel functions are used in this study, i.e., the linear kernel in SVM (Linear) and the RBF kernel in SVM (RBF).

\subsubsection{Linear Discriminant Analysis}
As another classification method, Linear Discriminant Analysis (LDA)~\cite{tharwat2016linear} assumes that the measurements from each class are normally distributed and the covariance of each class is identical. Compared with Principal Component Analysis (PCA), LDA focuses on maximizing the separability among known categories, instead of reducing the data dimension or finding the direction with the most variance. LDA accomplishes this objective by creating new axes for projecting the data points on to, with the distance between means of different groups maximized and the variance within each group minimized.

\subsubsection{Quadratic Discriminant Analysis}
Quadratic Discriminant Analysis (QDA)~\cite{tharwat2016linear} is similar to LDA. The difference is that QDA does not require the covariance between each class to be identical and the decision surface derived from the likelihood ratio test for classification is quadratic.

\subsubsection{Ridge}
Ridge Classifier (Ridge)~\cite{saunders1998ridge} is a modification of Ridge regression for classification tasks. Ridge regression modifies the objective function of regression model to prevent the over-fitting problem. By adding a penalty value to the sum of squared estimates of errors, Ridge regression can reduce the parameter values of those insignificant variables, as insignificant variables cannot reduce enough amount of sum of squared estimates of errors to compensate for the penalty.

Ridge classifier is essentially the same as ridge regression, but with some manipulation on the target variable. For the binary classification task, Ridge classifier is performed as follow. Firstly, the target variable is transformed into +1 or -1, based on the class to which it belongs. Secondly, a ridge regression model is built to predict the target variable. Thirdly, if the output is greater than 0, the data point is predicted as positive and vice versa. For the multi-class classification task, the general one-vs-the-rest approach can be performed.

\subsubsection{Decision Tree}
Decision Tree (DT)~\cite{breiman1984classification} is a structure composed of the root node, the internal node and the leaf node. A root node is the very first node for a decision tree. Leaf nodes are the last nodes of each branch which would determine the final prediction for all the data points that fall into this branch. Internal nodes are the intermediate nodes. For the root node and internal nodes, a single feature is used to separate the dataset into two categories, which can be numerical or categorical. To determine which feature to be chosen for each node, different metrics can be used, e.g., the Gini impurity or the information gain. The building process of a decision tree can be terminated in multiple conditions, e.g., when there is no feature left to use or the maximum tree depth is reached.

\subsubsection{Random Forest}
Random Forest (RF)~\cite{breiman2001random} uses different subsets of variables and observations to fit different decision trees and makes the final prediction based on the majority vote from individual trees. One bootstrapped sample, which is repetitively sampled from the training set, is used to fit one decision tree. For each node in a tree, $N$ distinct variables are sampled from the available feature set randomly and the best feature among selected is chosen for the node. The number $N$ is decided by the out-of-bag error, which is the error of the trained model to predict those data points that have never been included in any bootstrapped samples.

\subsubsection{Extra Trees}
Extra Trees (ET)~\cite{geurts2006extremely} is a modification of random forest. There are two major differences between extra trees and random forest. The first is that each tree is trained using the whole sample set in extra trees, rather than a bootstrapped sample subset as in random forest. The second is that the splitting in each tree is randomized in extra trees. A random value is selected for each feature, instead of picking the local optimal point for each feature based on criteria like Gini Impurity. Different random split schemes for each feature are evaluated and the one with the best score is chosen to split the node finally. It is shown that extra trees can generalize well and perform better on noisy data when compared with random forest.

\subsubsection{AdaBoost}
As another ensemble method, AdaBoost Classifier (AdaBoost)~\cite{freund1997decision} combines multiple weak learners to perform the classification task. These learners can be stamps that only consist of one root node and two leaves. Each of the learners has its own weight when voting for the final prediction, depending on how well it performs. Different from the bagging approach in random forests, AdaBoost allows the subsequent learner to learn from the mistakes of its previous learner through giving the mistakenly classified samples higher weights. This idea is named as boosting and has been used in other boosting-based classifiers.

\subsubsection{Gradient Boosting Classifier}
Gradient Boosting Classifier (GB)~\cite{friedman2001greedy} is another boosting-based ensemble model. The individual tree models are trained sequentially. After the first tree trained with the target, the residual can be calculated and the following tree would fit the residual of its previous one. In the classification case, the residual can be calculated as the difference between the predicted probability generated using the Logistic function and the true value, e.g., 0 for the negative class and 1 for the positive class. The final prediction can be made with a threshold, which would be 1 if the predicted probability is higher than the threshold and 0 otherwise. When there is no improvement with more trees or the maximum number of trees reaches the specified limit, the building process of GB would terminate.

\subsubsection{Extreme Gradient Boosting}
As an improvement from previous boosting models, Extreme Gradient Boosting (XGBoost)~\cite{chen2016xgboost} has been widely used in the data science community with its unique advantages. Similar with GB, XGBoost fits Classification and Regression Trees (CARTs) to the residuals. The data split scheme for each node of a tree is based on the similarity score. A regularization term can also be added to the calculation of similarity scores and output values, which shrinks both of these values and results in more pruning to prevent overfitting. XGBoost also contains many engineering improvements for the efficient training and inference.

\subsubsection{Light Gradient Boosting}
Light Gradient Boosting (LightGBM)~\cite{ke2017lightgbm} makes several improvements from XGBoost. Firstly, it uses the histogram algorithm, which converts a continuous variable into a discrete variable by splitting it into different bins and replacing all the data points that fall into the bin with an integer. This step could save a lot of memory and computation when iterating all the data samples. Secondly, it allows the leaf-wise tree growth instead of a level-wise one, which means that when building the tree, LightGBM looks for the leaf that can make the best progress and grows on that leaf. This is better because the leaf-wise tree growth method could focus on profitable leaves instead of growing all the leaves on the same level. LightGBM adds a maximum depth on trees to prevent overfitting. Thirdly, it uses Gradient-based One-Side Sampling (GOSS), which selects data points with the large gradient and randomly samples some data points with the low gradient to form a new data set. GOSS helps to balance data reduction and accuracy. Moreover, LightGBM applies Exclusive Feature Bundling (EFB) to bundle exclusive features together, which helps to reduce the feature dimension and increase the computation speed.

\subsubsection{CatBoost}
CatBoost Classifier (CatBoost)~\cite{prokhorenkova2018catboost} has three major improvements when compared to other gradient boosting models. Firstly, it builds symmetric or oblivious trees which use the same feature and splitting value for each node at the same level. This type of tree is simpler, faster to compute and capable of preventing overfitting. Tuning the hyper-parameters would not change the prediction accuracy significantly for this type of tree, and this specific tree model provides a good prediction in the first run. The second improvement is that CatBoost uses the categorical feature support, which transforms categorical variables into numeric variables using their label values. Greedy feature combinations are used in CatBoost, to prevent too many combinations. Finally, to solve the prediction shift problem, CatBoost utilizes the ordered boosting, which performs a random permutation of data samples.

\subsection{Deep Learning Models}
\subsubsection{Multilayer Perceptron}
\begin{figure*}[!htb]
\centering
\includegraphics[width=0.7\textwidth]{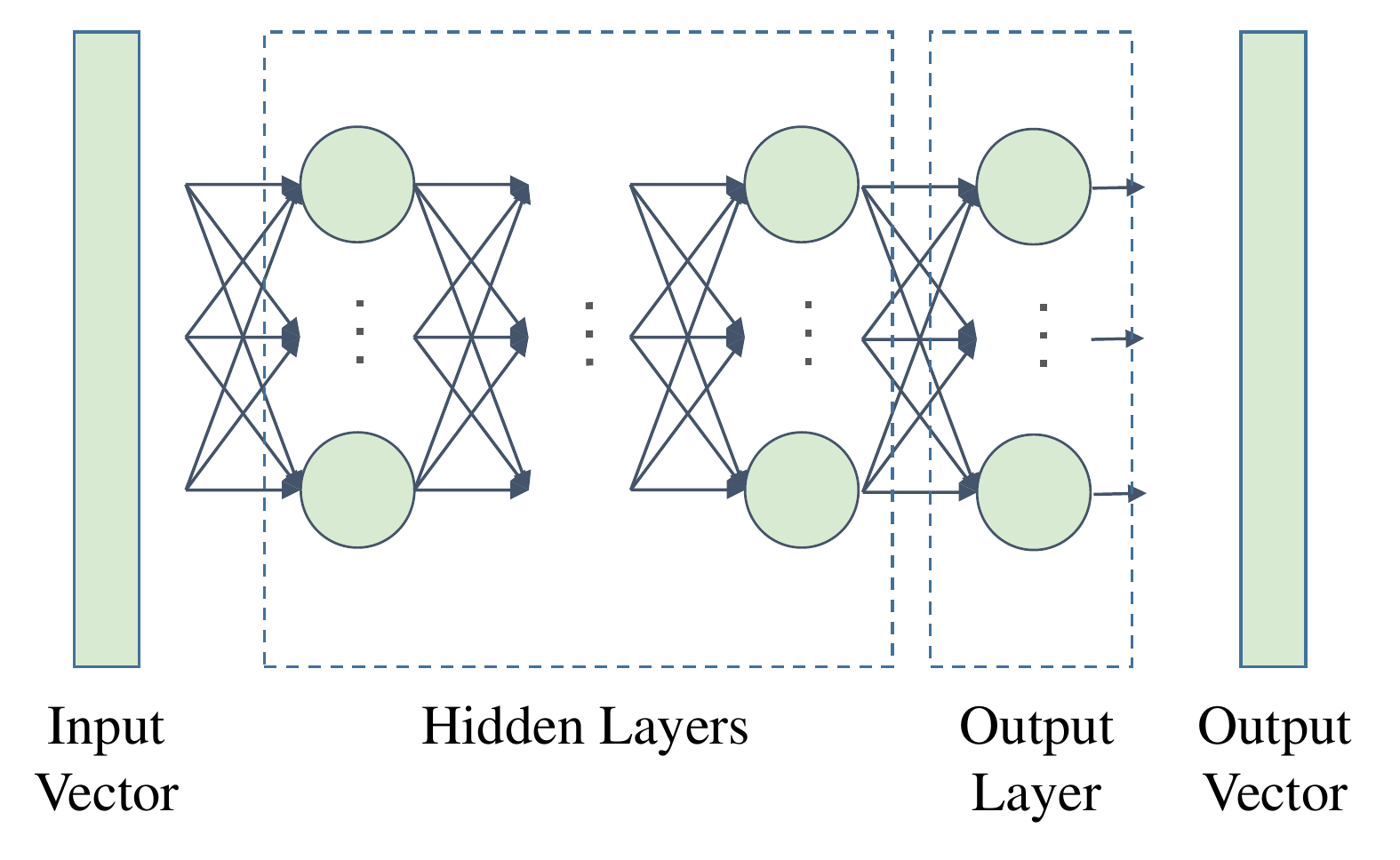}
\caption{The structure of MLP.}
\label{fig:MLP}
\end{figure*}

Multilayer Perceptron (MLP)~\cite{fawaz2019deep} is the simplest format of artificial neural networks, as shown in Figure~\ref{fig:MLP}. MLP uses hidden layers and nonlinear activation functions in a feed-forward neural network. In the neural network, a neuron is used to perform the basic operation. The mapping between input $\mathbf{x}$ and output $\mathbf{y}$ is as follows:
\begin{equation}
  y = \sigma(\mathbf{w} \cdot \mathbf{x} + b)
\end{equation}
\noindent where $\mathbf{w}$ and $b$ are parameters to learn from data, $\sigma(\cdot)$ is the nonlinear activation function.

The input features are formatted in a vector format, making MLP universal for many problems. However, this format fails to capture the temporal dependency or the spatial relationship in a two-dimensional image efficiently, thus is often outperformed by more efficient network structures, e.g., recurrent neural networks and convolutional neural networks.

\subsubsection{Recurrent Neural Networks}
\begin{figure*}[!htb]
\centering
\subfigure[][]{
  \includegraphics[width=0.7\textwidth]{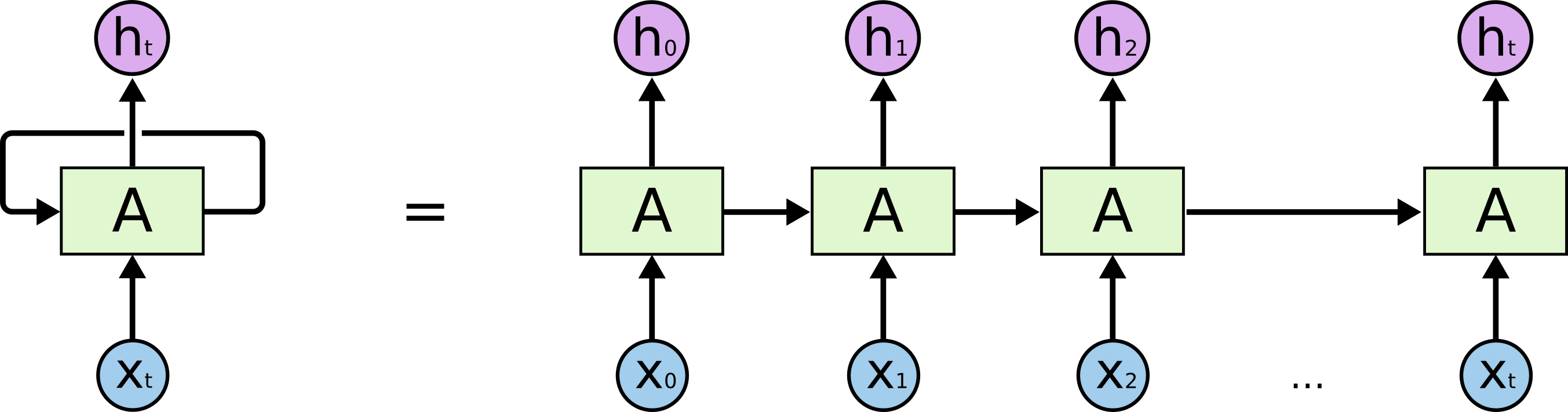}
  \label{fig:rnn-model}
} \\
\subfigure[][]{
  \includegraphics[width=0.45\textwidth]{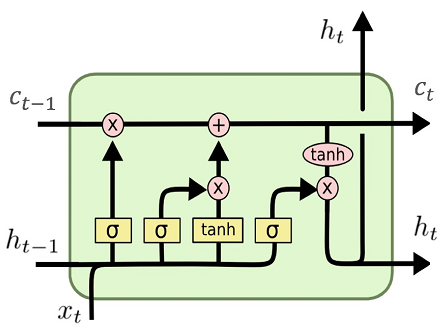}
  \label{fig:lstm-model}
}
\subfigure[][]{
  \includegraphics[width=0.45\textwidth]{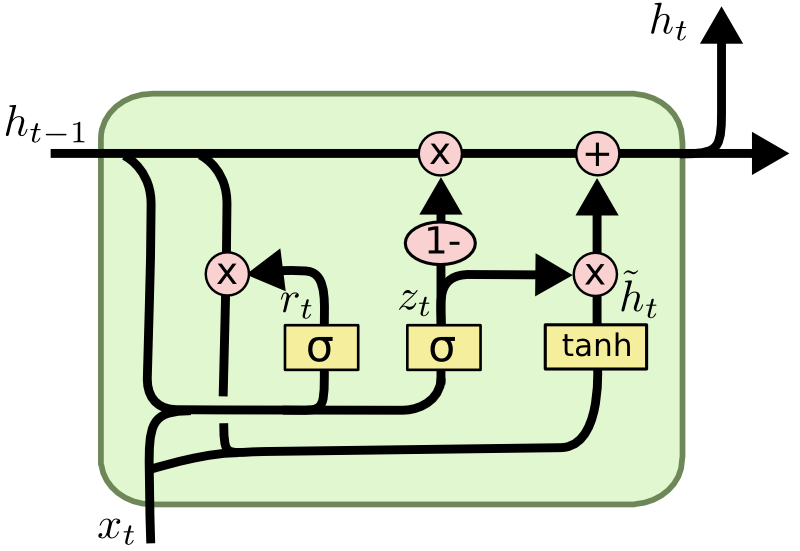}
  \label{fig:gru-model}
}
\caption[]{Recurrent neural networks~\cite{lstmlink}.
  \subref{fig:rnn-model} RNN;
  \subref{fig:lstm-model} LSTM;
  \subref{fig:gru-model} GRU.
}
\label{fig:rnns}
\end{figure*}

To capture the sequential dependency, Recurrent Neural Network (RNN)~\cite{rumelhart1986learning} adds previous outputs as inputs, while keeping hidden states. RNNs are useful for dealing with sequential data, e.g., natural language or time series. However, it is bothered by the vanishing gradient problem. Variants of RNNs are designed for alleviating this problem, e.g., Long Short-Term Memory (LSTM)~\cite{hochreiter1997long} and Gated Recurrent Unit (GRU)~\cite{cho2014learning}. The structures of RNNs are shown in Figure~\ref{fig:rnns}.

A typical LSTM cell~\cite{lstmlink} is shown in Figure~\ref{fig:lstm-model}, in which $t$ is the time slot index, $x_t$ is the input, i.e., a data sample, $h_t$ is the output value, and $c_t$ is the cell state. Compared with RNN, LSTM is featured with its three gate mechanisms, namely, the forget gate, the input gate, and the output gate.

The forget gate decides what information to throw away from the cell state:
\begin{equation}
f_t = \sigma(W_{xf} x_t + W_{hf} h_{t-1} + W_{cf} c_{t-1} + b_f)
\end{equation}
\noindent where $\sigma(*)$ denotes the sigmoid activation function, $W_{*}$ and $b_{*}$ represent parameters.

The input gate decides what new information is to be stored in the cell state:
\begin{equation}
\begin{split}
i_t &= \sigma(W_{xi} x_t + W_{hi} h_{t-1} + W_{ci} c_{t-1} + b_i) \\
\tilde{c}_t &= tanh(W_{xc} x_t + W_{hc} h_{t-1} + b_c) \\
c_t &= f_t * c_{t-1} + i_t * \tilde{c}_{t}
\end{split}
\end{equation}
\noindent where $tanh(*)$ denotes the tanh activation function, and $\tilde{c}_t$ denotes a matrix of new candidate values that could be added to the cell state.

The output gate decides what information to output from the cell state:
\begin{equation}
\begin{split}
o_t & = \sigma(W_{xo} x_t + W_{ho} h_{t-1} + W_{co} c_t + b_o) \\
h_t & = o_t * tanh(c_t)
\end{split}
\end{equation}

As another RNN variant, a typical GRU cell is shown in Figure~\ref{fig:gru-model}. Compared with LSTM, GRU combines the forget and input gates into the update gate. The cell state and hidden state are also merged, too. The other parts of LSTM are kept in GRU. Then the whole process of GRU can be shown as follows:
\begin{equation}
\begin{split}
z_t &= \sigma(W_{xz} x_t + W_{hz} h_{t-1} + b_z) \\
r_t &= \sigma(W_{xr} x_t + W_{hr} h_{t-1} + b_r) \\
\tilde{h}_t &= \sigma(W_{xo} x_t + W_{ho} r_t * h_{t-1} + b_o) \\
h_t &= (1 - z_t) * h_{t-1} + z_t * \tilde{h}_t
\end{split}
\end{equation}

\subsubsection{1D Convolutional Neural Networks}
Convolutional Neural Networks (CNNs) are extremely successful for two-dimensional data, but they can also be used for sequential data. In 1D Convolutional Neural Network (1D CNN), the convolutional kernel moves in one direction, instead of two directions. While recurrent neural networks are designed for sequential data, they are not as efficient as convolutional neural networks, which can be trained in parallel. Besides, many design ideas for two-dimensional CNN are transferred to 1D CNN, e.g., Fully Convolutional Network (FCN)~\cite{wang2017time} without using fully connected layers, Residual Neural Network (ResNet)~\cite{wang2017time} and Residual Convolutional Neural Network (ResCNN)~\cite{wang2017time} with the shortcut connection, InceptionTime~\cite{fawaz2020inceptiontime} inspired by the Inception structure, and XceptionTime~\cite{rahimian2020xceptiontime} inspired by the Xception structure.

As improvements over the normal 1D CNN, causal convolution and dilated convolution are proposed and used in Temporal Convolutional Networks (TCN)~\cite{bai2018empirical}. Causal convolution is proposed for dealing with sequences. Given the input sequence $\mathbf{X} = (x_1, x_2, \cdots, x_t)$ and the convolution filter (or kernel) $\mathbf{F} = (f_1, f_2, \cdots, f_K)$ with size $K$ in a hidden layer, the target is to predict the output sequence $\mathbf{Y} = (y_1, y_2, \cdots, y_t)$. Then the causal convolution for $x_t$ is
\begin{equation}
(\mathbf{F} * \mathbf{X})_{(x_t)} = \sum_{k=1}^K f_k x_{t-K+k}
\end{equation}
\noindent where only $x_i$ with $i \le t$ is used to avoid future information usage. With more hidden layers, each output element depends on a longer history of the input sequence.

\begin{figure*}[!htb]
\centering
\includegraphics[width=0.7\textwidth]{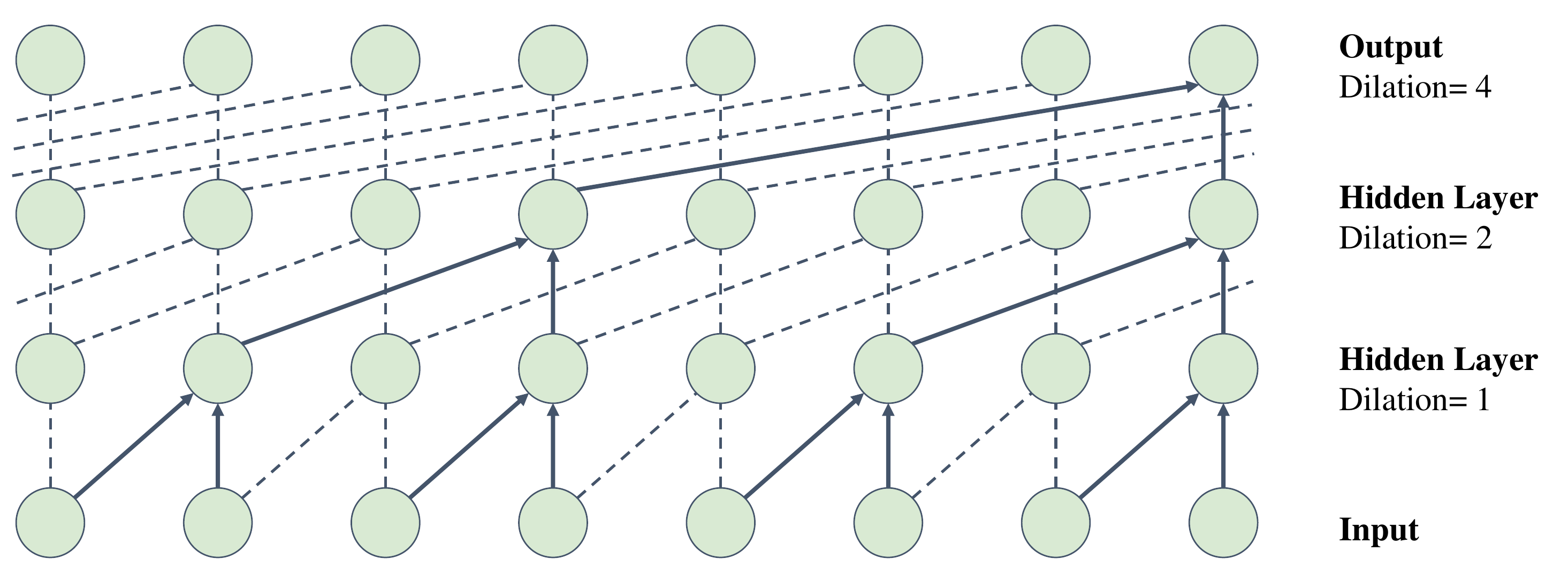}
\caption{The dilated causal convolution used in TCN.}
\label{fig:tcn}
\end{figure*}

Take a step further, one can add the dilatation rate $d$ in the convolution operation as follows:
\begin{equation}
(\mathbf{F} *_d \mathbf{X})_{(x_t)} = \sum_{k=1}^K f_k x_{t-(K-k)d}
\end{equation}
where a larger $d$ would increase the receptive field. In practice, $d$ is exponentially increased in different hidden layers, as shown in a specific example of dilated causal convolution in Figure~\ref{fig:tcn}, where $d=1, 2, 4$.

Furthermore, two recent 1D CNN variants are also Incorporated in this study, namely, Omni-Scale 1D-CNN (OS-CNN)~\cite{tang2020rethinking} and XCM~\cite{fauvel2020xcm}. OS-CNN can capture the proper kernel size during the model learning period, while XCM, an eXplainable Convolutional neural network for Multivariate time series classification, enables a good generalization ability and explainability.

\subsubsection{RNN-CNNs}
\begin{figure*}[!htb]
\centering
\includegraphics[width=0.7\textwidth]{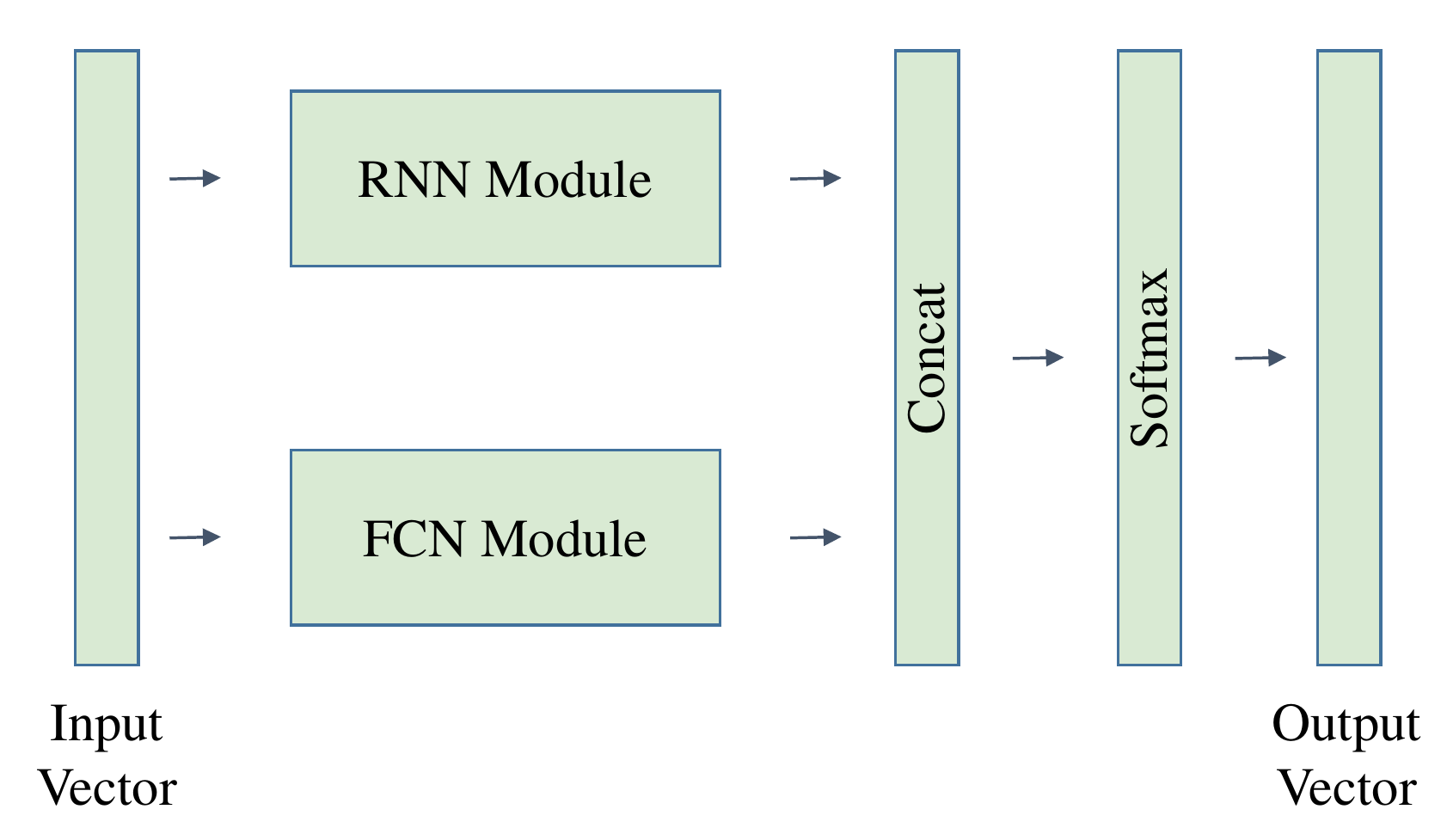}
\caption{The structure of RNN\_FCN.}
\label{fig:rnn-fcn}
\end{figure*}

To utilize the feature extraction abilities of both RNN and FCN, RNN\_FCN as well as its variants GRU\_FCN and LSTM\_FCN are proposed in~\cite{karim2017lstm} and further developed. The general structure of RNN\_FCN is shown in Figure~\ref{fig:rnn-fcn}, in which the outputs of the RNN module and FCN module are concatenated, before feeding into the softmax activation function that generates a probability distribution. In Figure~\ref{fig:rnn-fcn}, the RNN module can be replaced with GRU for GRU\_FCN and LSTM for LSTM\_FCN.

\subsubsection{Time Series Transformer}
\begin{figure*}[!htb]
\centering
\includegraphics[width=0.5\textwidth]{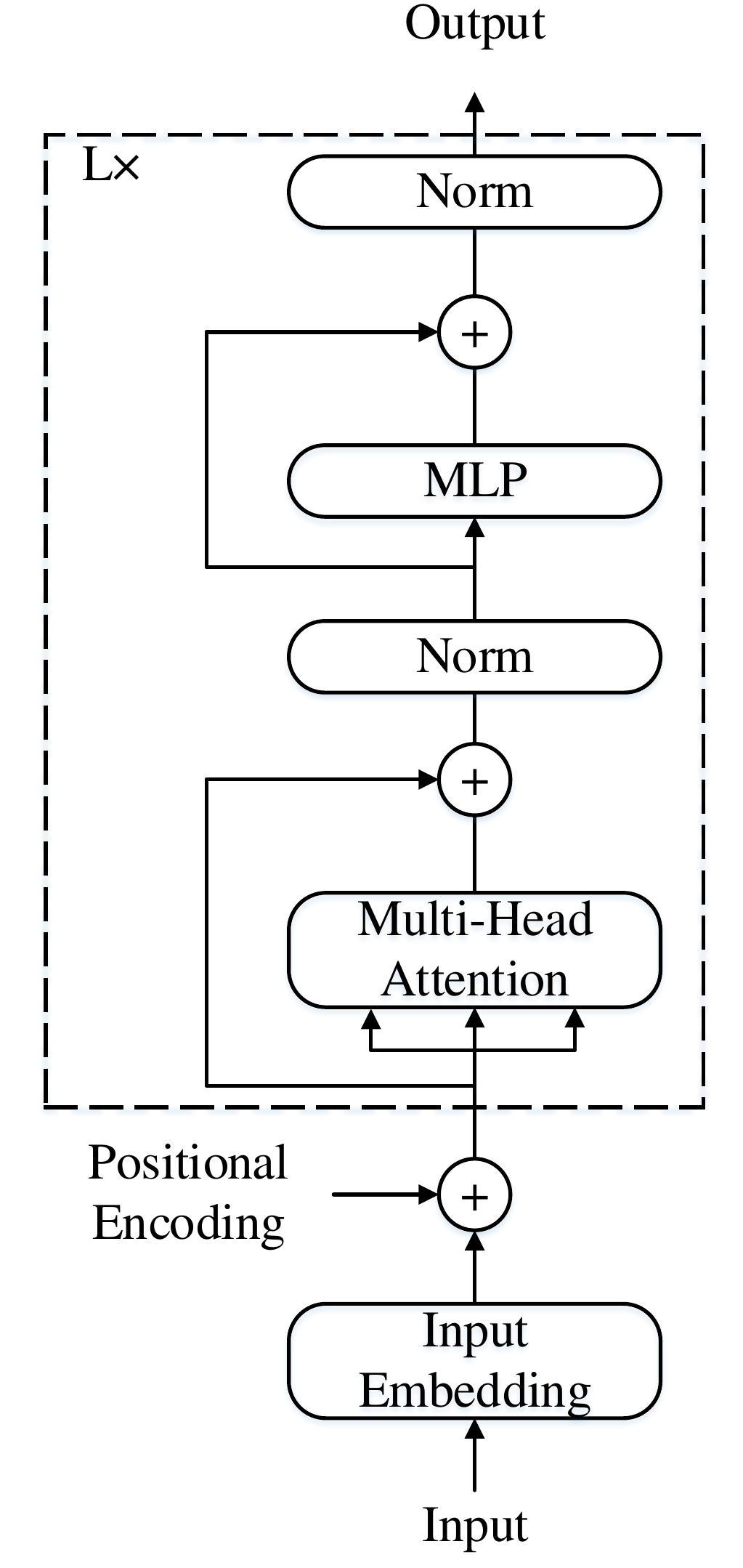}
\caption{The structure of Transformer Encoder.}
\label{fig:transformer}
\end{figure*}

The Transformer model was originally proposed by the Google team in 2017 and was adapted to machine translation~\cite{vaswani2017attention}. It abandoned the traditional recurrent neural network method of extracting sequence information and pioneered an attention mechanism to achieve fast parallelism, which improved the shortcomings of slow recurrent neural network training. Transformer is based on a typical encoder-decoder structure and only the encoder part is used in Time Series Transformer (TST)~\cite{zerveas2020transformer}, as shown in Figure~\ref{fig:transformer}.

The attention mechanism can be described as mapping a query and a set of key-value pairs to an output. The input consists of queries and keys of dimension $d_k$, and values of dimension $d_v$. The matrix of outputs is calculated as follows:
\begin{equation}
Attention(Q, K, V) = softmax(\frac{QK^T}{\sqrt{d_k}})V
\end{equation}
\noindent where $Q$ is the query matrix, $K$ is the key matrix, and $V$ is the value matrix. 

By using multiple attention heads, the multi-head attention is calculated as follows: 
\begin{equation}
MultiHead(Q, K, V) = Concat(head_1, \dots, head_h) W^O
\end{equation}
where $head_i = Attention(QW_i^Q, KW_i^K, VW_i^V)$, and $W$ are parameter matrices.

\section{Experiments}
\label{sec:experiments}
\subsection{Settings}
All the numerical experiments are conducted on a desktop computer. The main system configurations are as follows:
\begin{itemize}
    \item CPU: core i5-9600K CPU;
    \item GPU: NVIDIA GeForce RTX 2700 8GB;
    \item Memory: DDR4 32GB.
\end{itemize}

Python 3.7 is chosen as the programming language for implementing both the machine learning and deep learning models, because of its popularity and wide support in the machine learning community. Specifically speaking, the machine learning models are implemented with the packages including scikit-learn~\footnote{\url{https://scikit-learn.org/}}, xgboost~\footnote{\url{https://xgboost.readthedocs.io/}}, lightgbm~\footnote{\url{https://lightgbm.readthedocs.io/}}, and catboost~\footnote{\url{https://catboost.ai/}}. The deep learning models are implemented with the packages including pytorch~\footnote{\url{https://catboost.ai/}}, fastai~\footnote{\url{https://www.fast.ai/}}, and tsai~\footnote{\url{https://timeseriesai.github.io/tsai/}}. For tuning the hyper parameters on the validation set, pycaret~\footnote{\url{https://pycaret.org/}} is used for packing and tuning the machine learning models, and hyperopt~\footnote{\url{http://hyperopt.github.io/hyperopt/}} is used for searching the hyper parameter space of deep learning models. Due to space constraint, the full list of hyper parameter search space for all models is attached in Appendix A. More details of the experiment setting can be found in the notebooks contained in the associated Github repository~\footnote{\url{https://github.com/jwwthu/DL4Climate/tree/main/DroughtPrediction}}.

\subsection{Evaluation Metrics}
We use five evaluation metrics in this study to quantify the prediction performance of each model, namely, accuracy, precision, recall, F1 score and Matthews correlation coefficient (MCC). In the case of a binary classification, we have the following $2 \times 2$ confusion matrix to present the prediction results of a classifier in Table~\ref{tab:cm}.

\begin{table}[!htb]
\centering
\caption{The confusion matrix of a binary classifier.}
\label{tab:cm}
\begin{tabular}{|c|c|c|c|}
\hline
\multicolumn{2}{|c|}{} & \multicolumn{2}{c|}{Predication Outcome} \\
\cline{3-4}
\multicolumn{2}{|c|}{} & 1 & 0 \\
\hline
\multirow{2}{*}{Actual Value} & 1 & True Positive (TP) & False Negative (FN) \\
\cline{2-4}
& 0 & False Positive (FP) & True Negative (TN) \\
\hline
\end{tabular}
\end{table}

Then accuracy is defined as follows:
\begin{equation}
    accuracy = \frac{TP+TN}{TP+FN+FP+TN}
\end{equation}
\noindent where TP, TN, FN, and FP are defined in Table~\ref{tab:cm}.

Similarly, precision, recall, F1 score and MCC can be defined as follows:
\begin{equation}
    precision = \frac{TP}{TP+FP}
\end{equation}
\begin{equation}
    recall = \frac{TP}{TP+FN}
\end{equation}
\begin{equation}
    F1 = 2 \times \frac{precision \times recall}{precision+recall}
\end{equation}
\begin{equation}
    MCC = \frac{TP \times TN - FP \times FN}{\sqrt{(TP+FP)(TP+FN)(TN+FP)(TN+FN)}}
\end{equation}

The best value for accuracy, precision, recall and F1 score is 1, which represents a perfect prediction, and the worst value is 0. The best value for MCC is also 1, but the smallest value is -1, which represents a reverse prediction. The value of MMC being 0 represents the case of random guess, which is actually the worst case. In a reverse prediction, one can simply reverse the classifier’s outcome to get the ideal classifier.

Since we are dealing with a multi-class situation in this study, these metrics can be first calculated for each class independently, by taking the samples from a specific class as positive and the remaining samples as negative. Then the average value among all classes is calculated as the final result. Since this is an imbalanced dataset, the average value is calculated in a macro way without considering the proportion for each class in the dataset. Compared with using a weight that depends on the number of true labels of each class, the macro F1 score results in a bigger penalization when the model does not perform well with the minority classes. All the following results are the macro average values.

\subsection{Results}
The detailed results of machine learning models are listed in Table~\ref{tab:ml} and the detailed results of deep learning models are listed in Table~\ref{tab:dl}. We also calculate the best (max) value, the mean value and the standard deviation for each evaluation metrics, as shown in the last three rows of Table~\ref{tab:ml} and Table~\ref{tab:dl}.

From Table~\ref{tab:ml} and Table~\ref{tab:dl}, we have the following observations:
\begin{itemize}
\item The performance of deep learning models is closer to each other, compared with the case of machine learning models. This observation is based on the smaller standard deviation values in Table~\ref{tab:dl} and is reasonable because it is more common for some deep learning models sharing a similar mechanism, e.g., RNN, GRU and LSTM all belong to recurrent neural networks.
\item The deep learning models perform better than the machine learning models generally, as indicated by the higher mean values in Table~\ref{tab:dl}. This is also reasonable because of the strong learning ability of deep neural networks, which is also proven effective in relevant studies.
\item Among machine learning models, XGBoost achieves the highest accuracy and SVM with the RBF kernel achieves the highest F1 score and MCC. Among deep learning models, GRU achieves the highest accuracy, LSTM achieves the highest F1 score, and XceptionTime achieves the highest MCC. No single model can achieve the best performance for all evaluation metrics simultaneously.
\end{itemize}

\begin{table}[!htb]
\centering
\caption{The results of machine learning models.}
\label{tab:ml}
\begin{tabular}{llllll}
\hline
Model & Accuracy & Precision & Recall & F1 Score & MCC \\
\hline
LR & 0.715 & 0.272 & 0.248 & 0.255 & 0.270 \\
kNN & 0.564 & 0.217 & 0.216 & 0.215 & 0.097 \\
NB & 0.449 & 0.235 & 0.237 & 0.193 & 0.154 \\
SVM (linear) & 0.729 & 0.226 & 0.192 & 0.178 & 0.183 \\
SVM (RBF) & 0.617 & 0.283 & \textbf{0.308} & \textbf{0.284} & \textbf{0.298} \\
LDA & 0.720 & \textbf{0.285} & 0.254 & 0.259 & 0.258 \\
QDA & 0.726 & 0.180 & 0.167 & 0.141 & 0.014 \\
Ridge & 0.726 & 0.268 & 0.195 & 0.195 & 0.192 \\
DT & 0.724 & 0.150 & 0.173 & 0.153 & 0.075 \\
RF & 0.727 & 0.218 & 0.170 & 0.148 & 0.056 \\
ET & 0.727 & 0.158 & 0.167 & 0.141 & 0.013 \\
AdaBoost & 0.726 & 0.270 & 0.195 & 0.187 & 0.120 \\
GB & 0.726 & 0.281 & 0.223 & 0.232 & 0.230 \\
XGboost & \textbf{0.731} & 0.279 & 0.200 & 0.202 & 0.202 \\
LightGBM & 0.712 & 0.280 & 0.241 & 0.251 & 0.263 \\
CatBoost & 0.730 & 0.273 & 0.190 & 0.183 & 0.164 \\
\hline
Max & 0.731 & 0.285 & 0.308 & 0.284 & 0.298 \\
Mean & 0.691 & 0.242 & 0.211 & 0.201 & 0.162 \\
Std & 0.076 & 0.045 & 0.043 & 0.047 & 0.092 \\
\hline
\end{tabular}
\end{table}

\begin{table}[!htb]
\centering
\caption{The results of deep learning models.}
\label{tab:dl}
\begin{tabular}{llllll}
\hline
Model & Accuracy & Precision & Recall & F1 Score & MCC \\
\hline
MLP & 0.711 & 0.284 & 0.278 & 0.277 & 0.303 \\
RNN & 0.730 & 0.344 & 0.238 & 0.245 & 0.263 \\
GRU & \textbf{0.733} & 0.326 & 0.267 & 0.281 & 0.312 \\
LSTM & 0.720 & 0.327 & \textbf{0.294} & \textbf{0.302} & 0.320 \\
FCN & 0.711 & \textbf{0.345} & 0.283 & 0.291 & 0.314 \\
ResNet & 0.725 & 0.291 & 0.233 & 0.244 & 0.254 \\
ResCNN & 0.727 & 0.298 & 0.229 & 0.238 & 0.252 \\
TCN & 0.704 & 0.323 & 0.290 & 0.298 & 0.307 \\
InceptionTime & 0.730 & 0.316 & 0.253 & 0.261 & 0.300 \\
XceptionTime & 0.725 & 0.339 & 0.282 & 0.298 & \textbf{0.322} \\
OS-CNN & 0.730 & 0.301 & 0.248 & 0.261 & 0.283 \\
XCM & 0.710 & 0.320 & 0.273 & 0.286 & 0.296 \\
RNN\_FCN & 0.703 & 0.330 & 0.290 & 0.297 & 0.317 \\
GRU\_FCN & 0.721 & 0.308 & 0.269 & 0.281 & 0.294 \\
LSTM\_FCN & 0.724 & 0.317 & 0.264 & 0.277 & 0.295 \\
TST & 0.730 & 0.313 & 0.261 & 0.273 & 0.299 \\
\hline
Max & 0.733 & 0.345 & 0.294 & 0.302 & 0.322 \\
Mean & 0.721 & 0.318 & 0.266 & 0.276 & 0.296 \\
Std & 0.010 & 0.018 & 0.021 & 0.020 & 0.022 \\
\hline
\end{tabular}
\end{table}

\begin{figure*}[!htb]
\centering
\subfigure[][]{
    \includegraphics[width=0.45\textwidth]{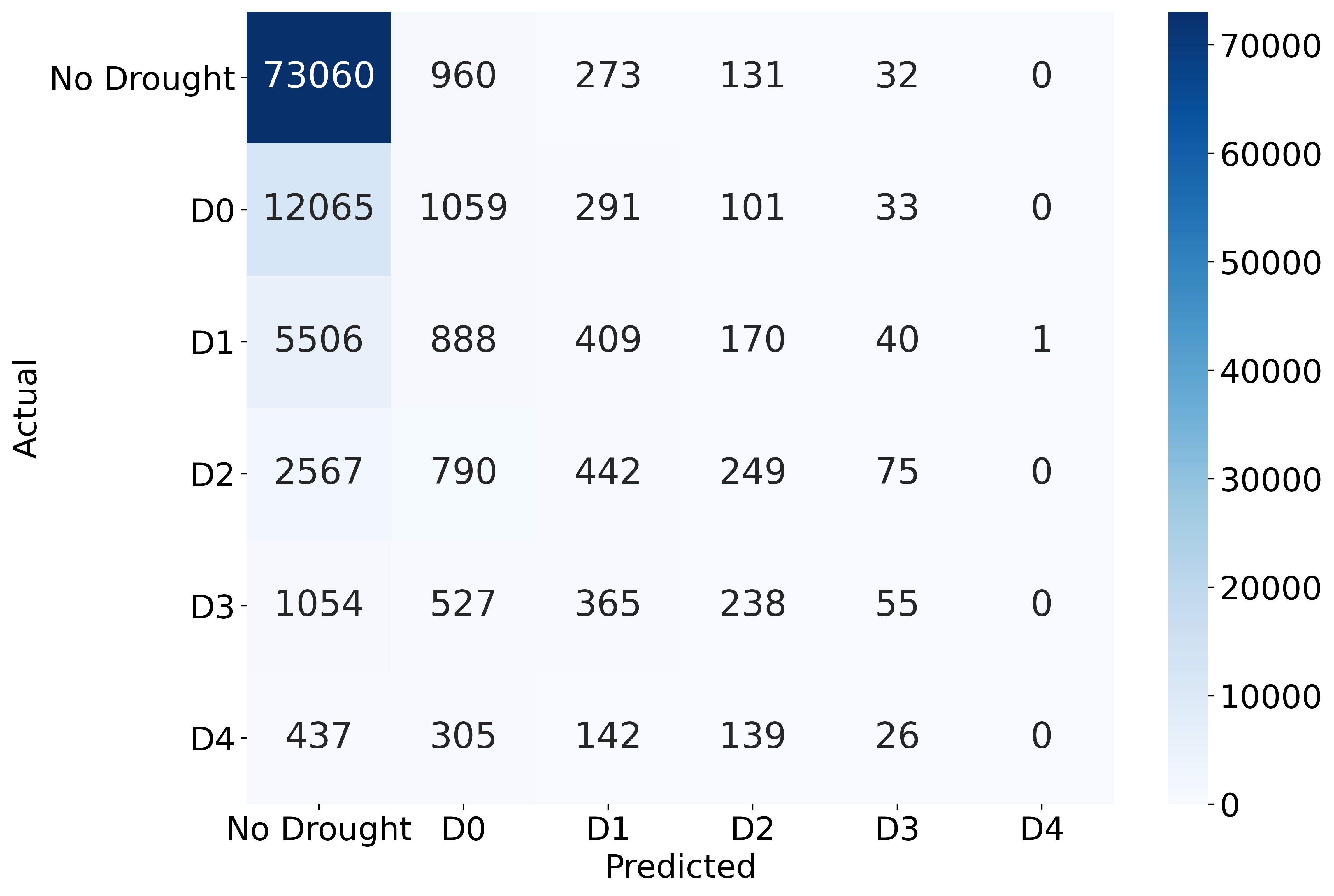}
    \label{fig:xgboost}
}
\subfigure[][]{
    \includegraphics[width=0.45\textwidth]{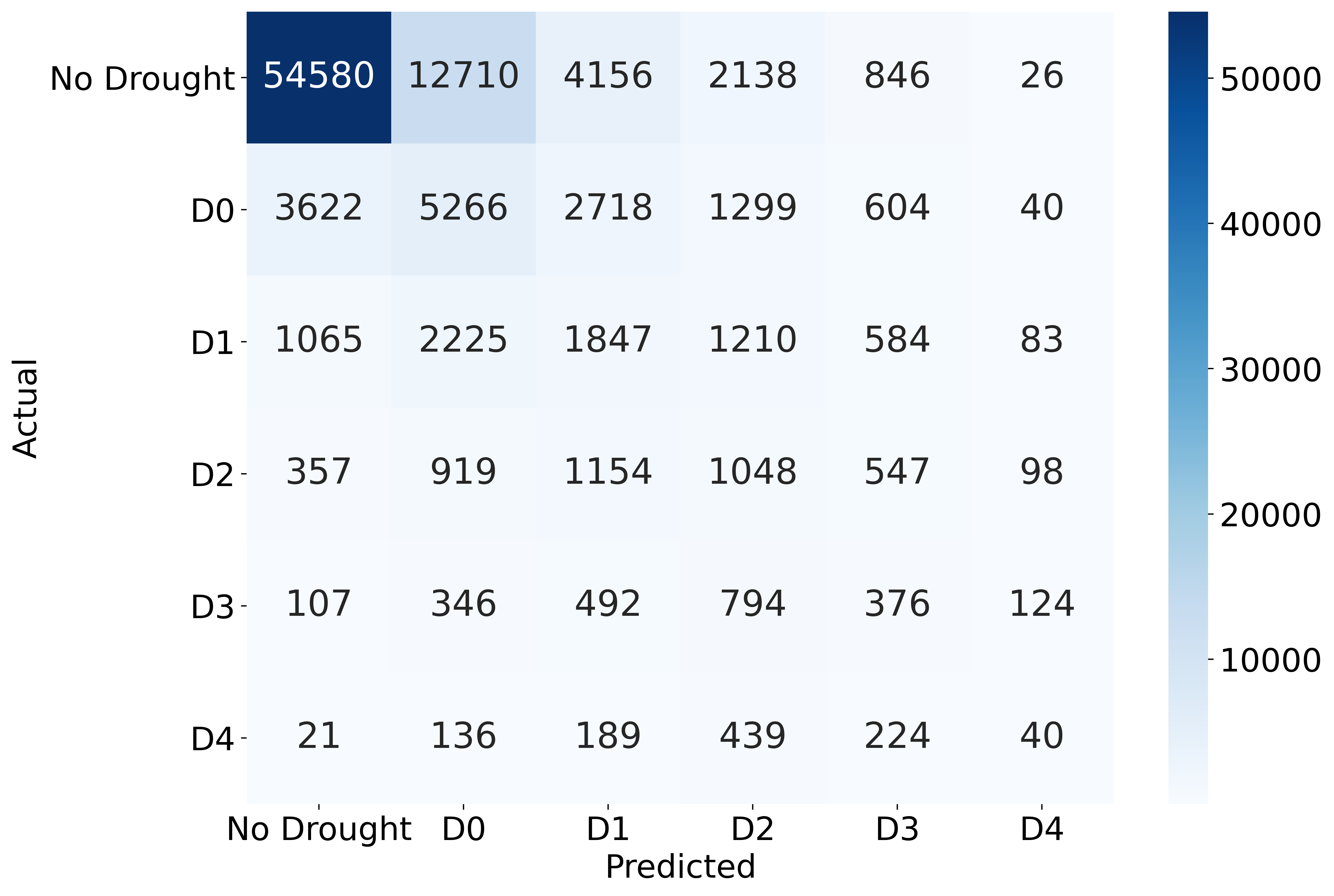}
    \label{fig:svm}
} \\
\subfigure[][]{
    \includegraphics[width=0.45\textwidth]{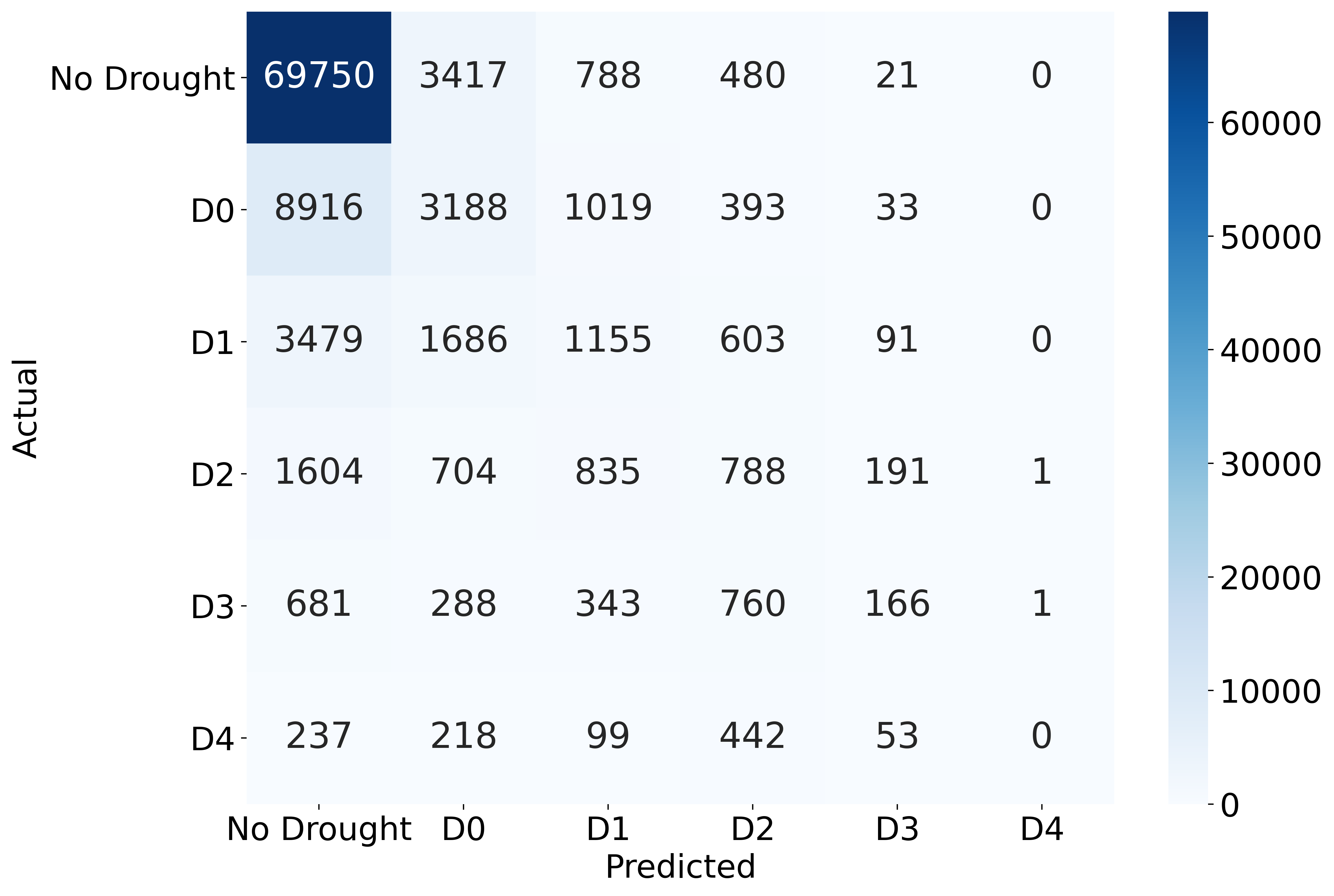}
    \label{fig:gru}
}
\subfigure[][]{
    \includegraphics[width=0.45\textwidth]{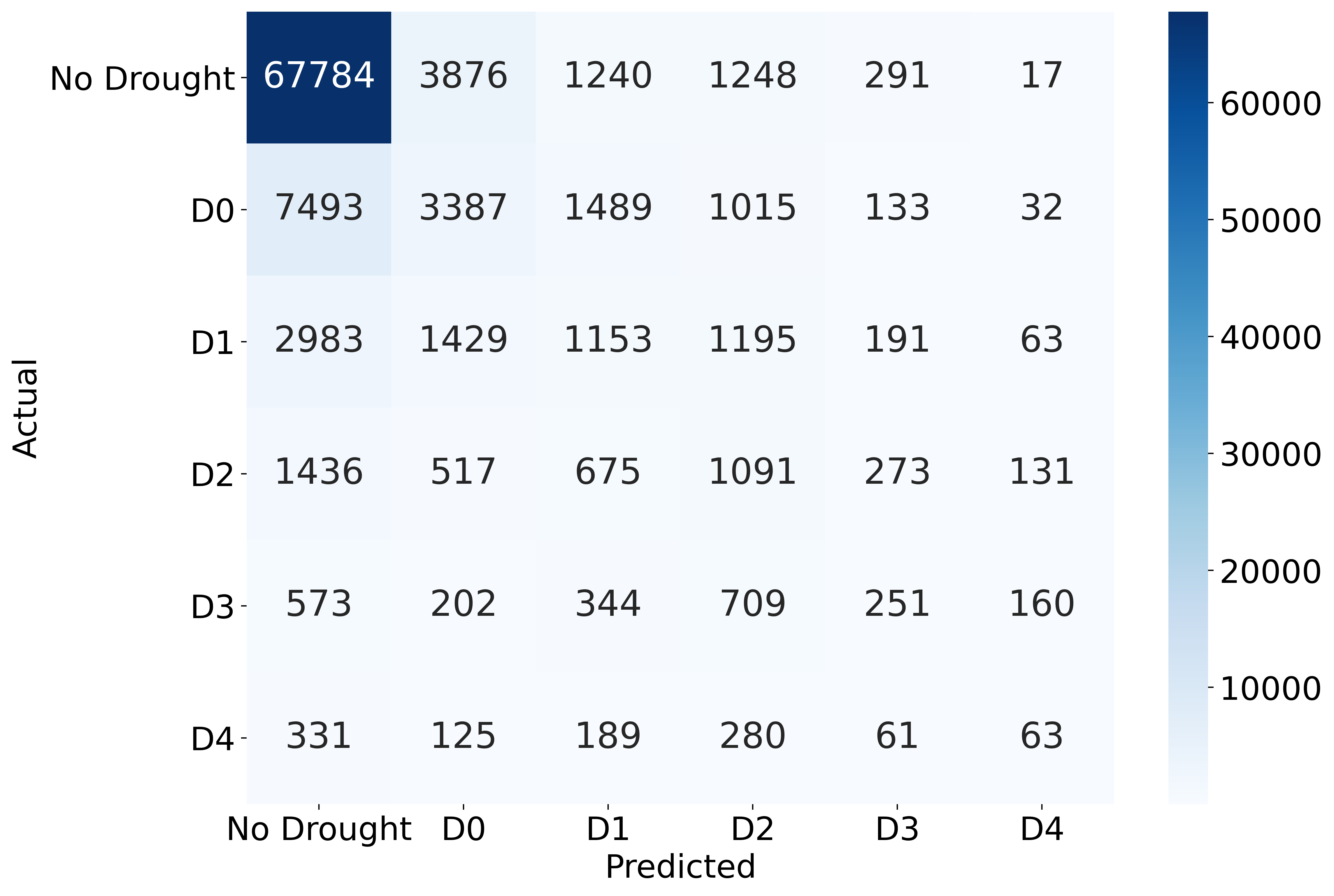}
    \label{fig:lstm}
}
\caption[]{Examples of confusion matrices.
    \subref{fig:xgboost} XGboost;
    \subref{fig:svm} SVM (rbf);
    \subref{fig:gru} GRU;
    \subref{fig:lstm} LSTM.
}
\label{fig:cms}
\end{figure*}

We further show the confusion matrices on the test set of XGBoost, SVM (RBF), GRU and LSTM in Figure~\ref{fig:cms} as the representatives. The confusion matrices for all the evaluated models on the test set are attached in Appendix B. It can be seen from Figure~\ref{fig:xgboost} and Figure~\ref{fig:gru}, the predictions of models with a higher accuracy are concentrated on the case of ``No Drought". The predictions of ``Exceptional Drought", i.e., ``D4", are rarely seen. However, for models with a higher macro F1 score in Figure~\ref{fig:svm} and Figure~\ref{fig:lstm}, the predictions are more balanced. However, there are still many errors when the prediction is ``D4" but the actual value is not, which degrades the macro F1 score by a large margin. The drought level prediction is still challenging, especially for predicting the rarely seen severe cases.

\section{Conclusion}
\label{sec:conclusion}
In this study, a comprehensive evaluation of state-of-the-art machine learning and deep learning models are conducted for drought prediction in this study, based on a real-world dataset collected in US spanning from 2000 to 2020. The results show that no single model can achieve the best performance for all five evaluation metrics used in this study simultaneously. Overall, deep learning models perform better than machine learning models. Among machine learning models, XGBoost and SVM (RBF) are relatively better and among deep learning models, GRU, LSTM, and XceptionTime are relatively better. Because of the imbalanced nature with the happening of different drought levels, the macro F1 score is far from satisfactory, even for the best model in this study and more exploration is needed in the future studies.

Two future directions are promising. The first direction is the combination of effective feature selection algorithms with machine learning models. For example, Principal Component Analysis (PCA) can be used to reduce the input dimension from 1710 (i.e., 90 days $\times$ 18 meteorological indicators) to a smaller value, e.g., 50 or 100. The reduced input features may not only accelerate the training process, but also reduces the possibility of overfitting, especially for deep learning models. However, the prediction performance may also be harmed. It is worthy to explore if there is an optimal tradeoff when the feature selection process is applied.

Another direction is the use of oversampling approach that creates synthetic minority class samples and helps to the imbalanced problem, e.g., Synthetic Minority Oversampling TEchnique (SMOTE). Contrary to feature selection, oversampling would add new training samples, increasing both the training time and computer memory requirements, which should be considered in practice. There may also be an optimal tradeoff between the extra training time consumption and the performance improvement, which is left for future studies.

\bibliography{mybibfile}

\begin{thebibliography}{10}
\expandafter\ifx\csname url\endcsname\relax
  \def\url#1{\texttt{#1}}\fi
\expandafter\ifx\csname urlprefix\endcsname\relax\def\urlprefix{URL }\fi
\expandafter\ifx\csname href\endcsname\relax
  \def\href#1#2{#2} \def\path#1{#1}\fi

\bibitem{tian2018agricultural}
Y.~Tian, Y.-P. Xu, G.~Wang, Agricultural drought prediction using climate
  indices based on support vector regression in xiangjiang river basin, Science
  of the Total Environment 622 (2018) 710--720.

\bibitem{dikshit2021improved}
A.~Dikshit, B.~Pradhan, A.~Huete, An improved spei drought forecasting approach
  using the long short-term memory neural network, Journal of environmental
  management 283 (2021) 111979.

\bibitem{mehr2021drought}
A.~D. Mehr, Drought classification using gradient boosting decision tree, Acta
  Geophysica (2021) 1--10.

\bibitem{mokhtar2021estimation}
A.~Mokhtar, M.~Jalali, H.~He, N.~Al-Ansari, A.~Elbeltagi, K.~Alsafadi, H.~G.
  Abdo, S.~S. Sammen, Y.~Gyasi-Agyei, J.~Rodrigo-Comino, Estimation of spei
  meteorological drought using machine learning algorithms, IEEE Access 9
  (2021) 65503--65523.

\bibitem{dikshit2021long}
A.~Dikshit, B.~Pradhan, A.~M. Alamri, Long lead time drought forecasting using
  lagged climate variables and a stacked long short-term memory model, Science
  of The Total Environment 755 (2021) 142638.

\bibitem{khan2020prediction}
N.~Khan, D.~Sachindra, S.~Shahid, K.~Ahmed, M.~S. Shiru, N.~Nawaz, Prediction
  of droughts over pakistan using machine learning algorithms, Advances in
  Water Resources 139 (2020) 103562.

\bibitem{zhang2019meteorological}
R.~Zhang, Z.-Y. Chen, L.-J. Xu, C.-Q. Ou, Meteorological drought forecasting
  based on a statistical model with machine learning techniques in shaanxi
  province, china, Science of the Total Environment 665 (2019) 338--346.

\bibitem{xu2018evaluation}
L.~Xu, N.~Chen, X.~Zhang, Z.~Chen, An evaluation of statistical, nmme and
  hybrid models for drought prediction in china, Journal of hydrology 566
  (2018) 235--249.

\bibitem{zhu2021internal}
S.~Zhu, Z.~Xu, X.~Luo, X.~Liu, R.~Wang, M.~Zhang, Z.~Huo, Internal and external
  coupling of gaussian mixture model and deep recurrent network for
  probabilistic drought forecasting, International Journal of Environmental
  Science and Technology 18~(5) (2021) 1221--1236.

\bibitem{banadkooki2021multi}
F.~B. Banadkooki, V.~P. Singh, M.~Ehteram, Multi-timescale drought prediction
  using new hybrid artificial neural network models, Natural Hazards 106~(3)
  (2021) 2461--2478.

\bibitem{malik2021prediction}
A.~Malik, Y.~Tikhamarine, S.~S. Sammen, S.~I. Abba, S.~Shahid, Prediction of
  meteorological drought by using hybrid support vector regression optimized
  with hho versus pso algorithms, Environmental Science and Pollution Research
  (2021) 1--20.

\bibitem{jiang2018geospatial}
W.~Jiang, L.~Zhang, Geospatial data to images: a deep-learning framework for
  traffic forecasting, Tsinghua Science and Technology 24~(1) (2018) 52--64.

\bibitem{jiang2020applications}
W.~Jiang, Applications of deep learning in stock market prediction: recent
  progress, Expert Systems with Applications.

\bibitem{jiang2020edge}
W.~Jiang, L.~Zhang, Edge-siamnet and edge-triplenet: new deep learning models
  for handwritten numeral recognition, IEICE Transactions on Information and
  Systems 103~(3) (2020) 720--723.

\bibitem{bishop2006pattern}
C.~M. Bishop, Pattern recognition and machine learning, springer, 2006.

\bibitem{altman1992introduction}
N.~S. Altman, An introduction to kernel and nearest-neighbor nonparametric
  regression, The American Statistician 46~(3) (1992) 175--185.

\bibitem{friedman1997bayesian}
N.~Friedman, D.~Geiger, M.~Goldszmidt, Bayesian network classifiers, Machine
  learning 29~(2) (1997) 131--163.

\bibitem{cortes1995support}
C.~Cortes, V.~Vapnik, Support-vector networks, Machine learning 20~(3) (1995)
  273--297.

\bibitem{tharwat2016linear}
A.~Tharwat, Linear vs. quadratic discriminant analysis classifier: a tutorial,
  International Journal of Applied Pattern Recognition 3~(2) (2016) 145--180.

\bibitem{saunders1998ridge}
C.~Saunders, A.~Gammerman, V.~Vovk, Ridge regression learning algorithm in dual
  variables, in: Proceedings of the Fifteenth International Conference on
  Machine Learning, 1998, pp. 515--521.

\bibitem{breiman1984classification}
L.~Breiman, J.~Friedman, C.~J. Stone, R.~A. Olshen, Classification and
  regression trees, CRC press, 1984.

\bibitem{breiman2001random}
L.~Breiman, Random forests, Machine learning 45~(1) (2001) 5--32.

\bibitem{geurts2006extremely}
P.~Geurts, D.~Ernst, L.~Wehenkel, Extremely randomized trees, Machine learning
  63~(1) (2006) 3--42.

\bibitem{freund1997decision}
Y.~Freund, R.~E. Schapire, A decision-theoretic generalization of on-line
  learning and an application to boosting, Journal of computer and system
  sciences 55~(1) (1997) 119--139.

\bibitem{friedman2001greedy}
J.~H. Friedman, Greedy function approximation: a gradient boosting machine,
  Annals of statistics (2001) 1189--1232.

\bibitem{chen2016xgboost}
T.~Chen, C.~Guestrin, Xgboost: A scalable tree boosting system, in: Proceedings
  of the 22nd acm sigkdd international conference on knowledge discovery and
  data mining, 2016, pp. 785--794.

\bibitem{ke2017lightgbm}
G.~Ke, Q.~Meng, T.~Finley, T.~Wang, W.~Chen, W.~Ma, Q.~Ye, T.-Y. Liu, Lightgbm:
  A highly efficient gradient boosting decision tree, Advances in neural
  information processing systems 30 (2017) 3146--3154.

\bibitem{prokhorenkova2018catboost}
L.~Prokhorenkova, G.~Gusev, A.~Vorobev, A.~V. Dorogush, A.~Gulin, Catboost:
  unbiased boosting with categorical features, in: Proceedings of the 32nd
  International Conference on Neural Information Processing Systems, 2018, pp.
  6639--6649.

\bibitem{fawaz2019deep}
H.~I. Fawaz, G.~Forestier, J.~Weber, L.~Idoumghar, P.-A. Muller, Deep learning
  for time series classification: a review, Data Mining and Knowledge Discovery
  33~(4) (2019) 917--963.

\bibitem{rumelhart1986learning}
D.~E. Rumelhart, G.~E. Hinton, R.~J. Williams, Learning representations by
  back-propagating errors, nature 323~(6088) (1986) 533--536.

\bibitem{cho2014learning}
K.~Cho, B.~Van~Merri{\"e}nboer, C.~Gulcehre, D.~Bahdanau, F.~Bougares,
  H.~Schwenk, Y.~Bengio, Learning phrase representations using rnn
  encoder-decoder for statistical machine translation, arXiv preprint
  arXiv:1406.1078.

\bibitem{hochreiter1997long}
S.~Hochreiter, J.~Schmidhuber, Long short-term memory, Neural computation 9~(8)
  (1997) 1735--1780.

\bibitem{wang2017time}
Z.~Wang, W.~Yan, T.~Oates, Time series classification from scratch with deep
  neural networks: A strong baseline, in: 2017 International joint conference
  on neural networks (IJCNN), IEEE, 2017, pp. 1578--1585.

\bibitem{bai2018empirical}
S.~Bai, J.~Z. Kolter, V.~Koltun, An empirical evaluation of generic
  convolutional and recurrent networks for sequence modeling, arXiv preprint
  arXiv:1803.01271.

\bibitem{fawaz2020inceptiontime}
H.~I. Fawaz, B.~Lucas, G.~Forestier, C.~Pelletier, D.~F. Schmidt, J.~Weber,
  G.~I. Webb, L.~Idoumghar, P.-A. Muller, F.~Petitjean, Inceptiontime: Finding
  alexnet for time series classification, Data Mining and Knowledge Discovery
  34~(6) (2020) 1936--1962.

\bibitem{rahimian2020xceptiontime}
E.~Rahimian, S.~Zabihi, S.~F. Atashzar, A.~Asif, A.~Mohammadi, Xceptiontime:
  independent time-window xceptiontime architecture for hand gesture
  classification, in: ICASSP 2020-2020 IEEE International Conference on
  Acoustics, Speech and Signal Processing (ICASSP), IEEE, 2020, pp. 1304--1308.

\bibitem{tang2020rethinking}
W.~Tang, G.~Long, L.~Liu, T.~Zhou, J.~Jiang, M.~Blumenstein, Rethinking 1d-cnn
  for time series classification: A stronger baseline, arXiv preprint
  arXiv:2002.10061.

\bibitem{fauvel2020xcm}
K.~Fauvel, T.~Lin, V.~Masson, {\'E}.~Fromont, A.~Termier, Xcm: An explainable
  convolutional neural network for multivariate time series classification,
  arXiv preprint arXiv:2009.04796.

\bibitem{karim2017lstm}
F.~Karim, S.~Majumdar, H.~Darabi, S.~Chen, Lstm fully convolutional networks
  for time series classification, IEEE access 6 (2017) 1662--1669.

\bibitem{zerveas2020transformer}
G.~Zerveas, S.~Jayaraman, D.~Patel, A.~Bhamidipaty, C.~Eickhoff, A
  transformer-based framework for multivariate time series representation
  learning, arXiv preprint arXiv:2010.02803.

\bibitem{park2019prediction}
H.~Park, K.~Kim, et~al., Prediction of severe drought area based on random
  forest: Using satellite image and topography data, Water 11~(4) (2019) 705.

\bibitem{hao2018seasonal}
Z.~Hao, V.~P. Singh, Y.~Xia, Seasonal drought prediction: advances, challenges,
  and future prospects, Reviews of Geophysics 56~(1) (2018) 108--141.

\bibitem{vidyarthi2020knowledge}
V.~K. Vidyarthi, A.~Jain, Knowledge extraction from trained ann drought
  classification model, Journal of Hydrology 585 (2020) 124804.

\bibitem{lstmlink}
Understanding lstm networks [online], Available:
  \url{https://colah.github.io/posts/2015-08-Understanding-LSTMs/}, accessed
  on: May 22, 2021 (2015).

\bibitem{vaswani2017attention}
A.~Vaswani, N.~Shazeer, N.~Parmar, J.~Uszkoreit, L.~Jones, A.~N. Gomez,
  L.~Kaiser, I.~Polosukhin, Attention is all you need, arXiv preprint
  arXiv:1706.03762.

\end{thebibliography}

\end{document}